\documentclass{article}
\usepackage{spconf,amsmath,graphicx}

\usepackage{acronym}

\usepackage{amsmath,graphicx}
\usepackage{todonotes}

\def\T{\mathrm T}

\def\xvec{\mathbf x}

\def\yvec{\mathbf y}

\def\Xmat{\mathbf X}

\def\Ymat{\mathbf Y}

\usepackage{lineno,hyperref}
\usepackage{cite}
\usepackage{amsmath,graphicx}

\usepackage{amssymb,amsmath,comment,caption, subcaption, graphicx,amsmath,mathrsfs, array, hyperref, multirow}
\usepackage{booktabs}
\usepackage{diagbox}
\usepackage{xcolor}
\usepackage{float}
\modulolinenumbers[5]

\title{Beyond Isolated Utterances: Conversational Emotion Recognition}

\name{Raghavendra Pappagari$^1$, Piotr \.Zelasko$^{1,2}$, Jes\'us Villalba$^{1,2}$, Laureano Moro-Velazquez$^1$, Najim Dehak$^{1,2}$}
\address{$^1$Center for Language and Speech Processing, $^2$Human Language Technology Center of Excellence,
\\Johns Hopkins University, Baltimore, MD, USA \\
\tt \{rpappag1, pzelasko, jvillal7, laureano, ndehak3\}@jhu.edu}

\begin{document}

\maketitle
\begin{abstract}
Speech emotion recognition is the task of recognizing the speaker's emotional state given a recording of their utterance.
While most of the current approaches focus on inferring emotion from isolated utterances, we argue that this is not sufficient to achieve conversational emotion recognition (CER) which deals with recognizing emotions in conversations.
In this work, we propose several approaches for CER by treating it as a sequence labeling task.
We investigated transformer architecture for CER and, compared it with ResNet-34 and BiLSTM architectures in both contextual and context-less scenarios using IEMOCAP corpus.
Based on the inner workings of the self-attention mechanism, we proposed \textit{DiverseCatAugment} (\textit{DCA}), an augmentation scheme, which improved the transformer model performance by an absolute 3.3\% micro-f1 on conversations and 3.6\% on isolated utterances.
We further enhanced the performance by introducing an interlocutor-aware transformer model where we learn a dictionary of interlocutor index embeddings to exploit diarized conversations.

\end{abstract}


\section{Introduction}

Speech emotion recognition is the task of recognizing the speaker's emotional state from their speech.
The speaker's emotion might be affected by several factors such as demographics~\cite{koolagudi2012emotion}, age~\cite{paulmann2008aging}, and conversational context.
Recognizing the speaker's emotion is usually performed on the utterance level, i.e., the model predicts one emotion class for the whole utterance~\cite{huang2014speech, lim2016speech, cho2018deep, trigeorgis2016adieu, pappagari2020copypaste,feng2020end, lu2020speech}.
The core assumption of this approach is that isolated utterances, generally less than 10s in duration, contain only one speaker and one emotion throughout the utterance.
These approaches can be applied to conversational speech, provided that an utterance-level segmentation is available either from another system or a human annotator and assuming the utterance contains only one emotion.

Recently focus has shifted towards conversational emotion recognition (CER)~\cite{hazarika2018conversational, majumder2019dialoguernn, li2020hitrans, zhang2019modeling, grimm2007primitives, metallinou2013tracking, eyben2010line, schmitt2019continuous} to predict emotion more accurately using context.
One study explores a fixed context (4 recent utterances) and speaker-specific modeling using gated recurrent unit (GRU) architecture~\cite{hazarika2018conversational}. 
Their model, referred to as conversational memory network, uses attention mechanism and memory hopping to combine information from multiple streams of representations and to attend to history. 
The main limitation of this approach is its fixed context and a lack of extensibility to multi-party conversations.
Model proposed in~\cite{majumder2019dialoguernn}, referred to as DialogueRNN, overcomes limitations of the fixed context and also proposes to use separate GRUs to model speaker, emotion and global context.
Authors in~\cite{zhang2019modeling} propose to use graph based neural net by defining utterances and speakers as nodes to exploit context and speaker dependencies.
A transformer model with pairwise speaker verification as auxiliary task is proposed in~\cite{li2020hitrans} to encode context and speaker information into the model hidden representations.
Even though the above approaches provide good CER performance, all of them are fundamentally limited by their reliance on the availability of a segmentation of the recording/transcript, and their strong assumptions about each speaker turn consisting of just a single emotion.

\begin{figure*}
    \centering
    \includegraphics[width=0.8\textwidth, scale=0.5]{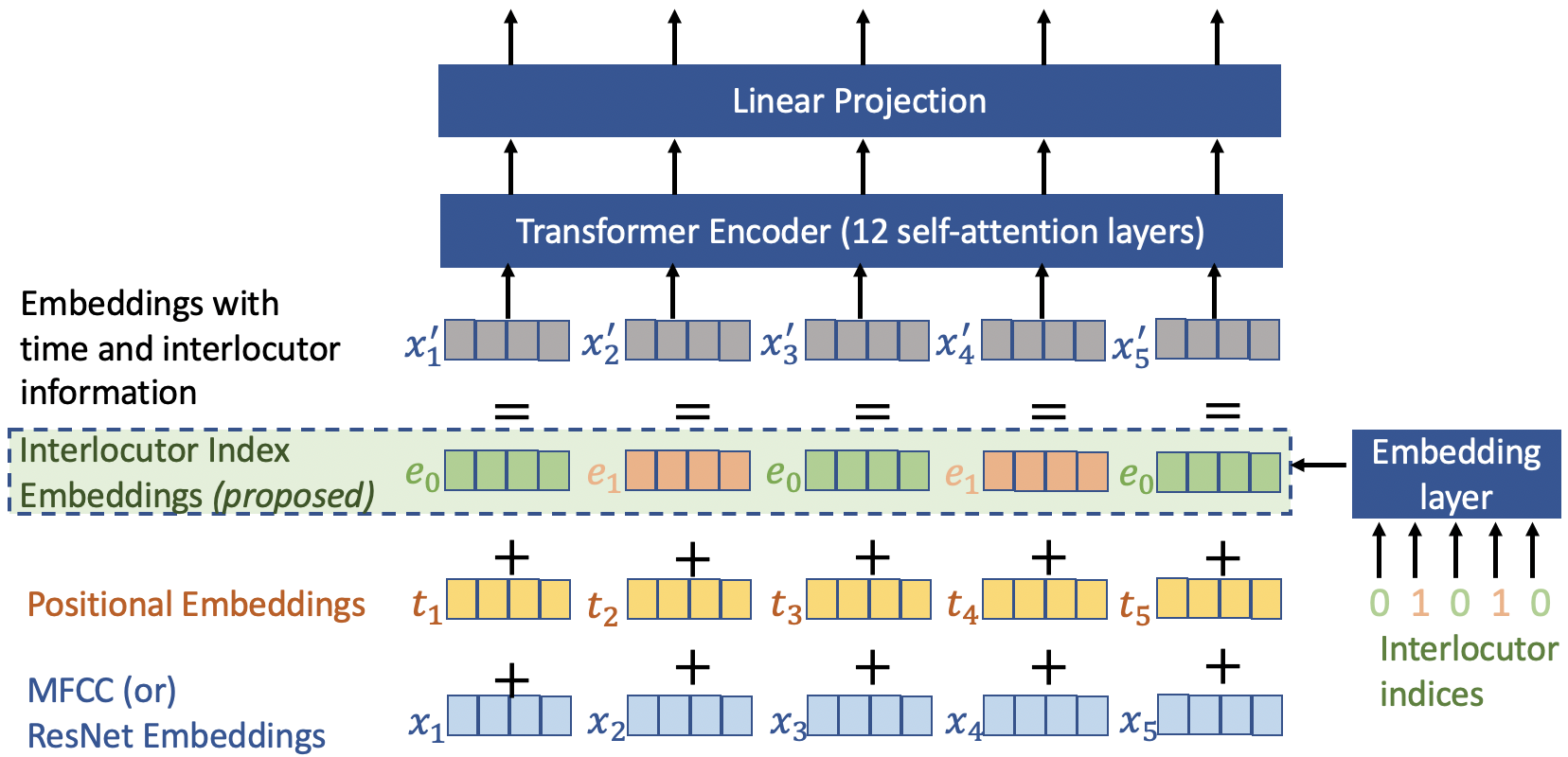}
    \caption{Transformer block diagram. Interlocutor index embeddings are used only with ResNet embeddings input in ResNet+Transformer model}
    \label{fig:transformer_block_diagram}
\end{figure*}

CER without requiring segmented recordings is explored in~\cite{grimm2007primitives,metallinou2013tracking, eyben2010line, schmitt2019continuous} by predicting emotional attributes on a frame-level.
Authors in~\cite{grimm2007primitives} use a fuzzy logic estimator while~\cite{metallinou2013tracking} propose an optimal statistical mapping between audiovisual features and emotion attributes using Gaussian mixture model (GMM).
Deep learning models such as CNN and LSTM are used in~\cite{eyben2010line, schmitt2019continuous} for frame-level prediction.

In this work, we present transformer-based models for CER by treating it as a sequence labeling task, where short duration frames of the speech signal are assigned emotion labels by a model that looks at the broader context.
Based on self-attention operation, we proposed \textit{DiverseCatAugment} (\textit{DCA}), an augmentation scheme to improve transformer model performance.
We quantified the effect of context by comparing models trained on isolated utterances and conversations.
We compared transformer architecture with several neural architectures: ResNet-34, which models context locally in each layer and globally with stacked layers and; BiLSTM, which captures context sequential manner.
To leverage both the local and global context modeling strengths of ResNet-34 and transformer architectures, we explore their joint training.
The resulting model is further enhanced with interlocutor index embeddings -- a novel input that is designed to make the model better aware of interlocutors in a conversation.
The proposed models can deal with multi-party conversations and do not assume one emotion per turn.

The rest of the paper is organized as follows. 
First, we present our models in Section~\ref{sec:approach} and the proposed \textit{DCA} augmentation scheme in Section~\ref{sec:DCA}.
Then, the experimental setup and results are detailed in Section~\ref{sec:experimental_setup} and~\ref{sec:results} respectively.
Finally, conclusions of our work and future directions are discussed in Section~\ref{sec:conclusion}.

\section{Our Models}
\label{sec:approach}

We present transformer-based models that predict emotion on a frame-level.
Our experiments include the use of a basic transformer architecture and also a combination of transformer and a CNN architecture. 
For all the approaches in this work, we employ MFCC features as input, with 25 ms frame length and 10 ms shift.

\textit{Baseline models:} The proposed transformer models are compared with two baseline models employed in previous studies: one using a BiLSTM architecture~\cite{lee2015high} and the other employing CNN~\cite{pappagari2020copypaste}. 
We use BiLSTM and CNN models as the mechanism of exploiting context is different in them compared to the transformer. 
BiLSTM learns context information in a sequential manner; CNN exploits local context in each layer and global context with a stack of layers; in contrast, the transformer has access to the entire conversational context in every self-attention layer.
Also, self-attention operation in the transformer model allows attending other frames with the same emotion in the sequence while convolutional operation treats all frames inside a receptive field in a similar manner disregarding their class label.
Our BiLSTM architecture contains a sequence of 6 bi-directional LSTM layers followed by a dropout layer and 2 fully connected layers to obtain logits. 
For CNN, we use ResNet-34 model architecture reported in~\cite{pappagari2020x} without pooling layer.

\subsection{Transformer model}
\label{subsec:transformer_model}

As the employed frame length is very small to predict an emotion, the context plays a crucial role in deciding the emotion of a frame.
Hence, an architecture that can use context efficiently is crucial.
Recently, transformer architecture has shown to outperform other neural architectures in several speech and NLP tasks~\cite{vaswani2017attention,devlin2019bert,wolf2020transformers,karita2019comparative}.
It contains a sequence of self-attention operations which are designed to exploit long-range dependencies in the input sequence.
Our architecture contains a sequence of 12 self-attention layers as in the standard BERT base model~\cite{devlin2019bert}.
A schematic of the transformer model is shown in Fig.~\ref{fig:transformer_block_diagram}.
For this model, we use MFCC features as input. 
As the entire input sequence is processed simultaneously in self-attention layers, the order of the input sequence does not matter.
However, the sequence information could be useful for the CER task.
We encode position information by learning a set of positional embeddings during training and adding them to the input sequence.
For training, we use an input sequence length of 2048 hence we learn 2048 positional embeddings during training.

\subsection{ResNet+Transformer model}

Several studies suggest that down-sampling input representation using convolutional layers before processing with transformer layers provides better results for ASR~\cite{lu2020exploring,mohamed2019transformers}.
Intuitively, convolutional layers use local context to produce better contextual features.
In this work, we used a pre-trained ResNet-34 to process input MFCC features and fed its output to the transformer layers.
ResNet-34 is pre-trained on the speaker classification task.
We jointly trained ResNet-34 and transformer to exploit the benefits of both transfer learning and transformer model capabilities.

\subsection{Interlocutor-aware ResNet+Transformer model}

A conversation is structured as a sequence of turns by all participating speakers.
The emotion of a speaker in each turn could depend on that speaker's emotions in previous turns and also on the interlocutor's emotions~\cite{hatfield1993emotional, smirnov2019emotions}.
Hence, we expect the model to perform better when the model knows who is speaking when in the conversation.

We propose to infuse interlocutor information by learning an interlocutor index embedding and adding it to the ResNet embeddings along with positional embeddings.
This method assumes the availability of speaker segmented conversations.
Our model schematic with interlocutor index embeddings is shown in Fig.~\ref{fig:transformer_block_diagram}.
During training, we indexed the speakers in a conversation and represented them with one-hot encoding.
Indices are assigned following the order in which the interlocutors appear in the conversation.
We passed the one-hot encoding through an embedding layer to get interlocutor index embedding.
The embedding layer learns a dictionary of embeddings which acts as a lookup table.
The dictionary size is set to the maximum number of speakers that can appear in a training sample. 
Then, we added interlocutor index embeddings to the ResNet output to incorporate interlocutor information into the transformer layers.
With this additional information, we expect the model learns to distinguish the speakers in the conversation and make accurate emotion predictions.
Note that this approach does not assume known speakers at the test time, it only assumes speaker diarized recording.
This model can deal with multi-party conversations but during test time it is limited by the maximum number of speakers seen in a training conversation.

\section{Diverse Category Augment Scheme}
\label{sec:DCA}

In this section, we present a data augmentation scheme, named as \textit{DiverseCatAugment} (\textit{DCA}), motivated by the inner workings of the self-attention operation.
Given an input sequence of vectors $\Xmat=[\xvec_{1}, \xvec_{2}, ..., \xvec_{N}]$, we perform self-attention operation and obtain a sequence of vectors $\Ymat=[\yvec_{1}, \yvec_{2}, ..., \yvec_{N}]$.
Self-attention operation (dot-product variation) as shown in~\eqref{eq:self_attention} includes finding dot-product between every vector in the sequence i.e., $\Xmat \cdot \Xmat^\T$. 
On the dot-product matrix, the softmax operation is employed to obtain normalized similarities for each vector with other vectors in the sequence.
Then, the dot product matrix is multiplied with the input sequence $\Xmat$ to obtain $\Ymat$. 
In essence, every vector in $\Ymat$ is a weighted sum of vectors in $\Xmat$, as shown in~\eqref{eq:weighted_sum} with weights being the normalized similarities with other vectors in the sequence as shown in~\eqref{eq:softmax_operation}.

\begin{equation*}
    \mathrm{Let}\, \Xmat=[\xvec_{1}, \xvec_{2}, ..., \xvec_{N}],\, \Ymat=[\yvec_{1}, \yvec_{2}, ..., \yvec_{N}]
\end{equation*}

\begin{center}
    where $\xvec_i, \yvec_i\in\mathbb{R}^d, \, \forall i\in\mathcal[1, N]$
\end{center}
\begin{equation}
    \Ymat=\mathrm{softmax}(\frac{\Xmat \cdot \Xmat^\T}{\sqrt{d}})\, \Xmat
    \label{eq:self_attention}
\end{equation}
\begin{equation}
    \yvec_{i}=\sum_{j=1}^{N} w_{ij} \xvec_{j}, \forall i\in\mathcal[1, N]  \label{eq:weighted_sum}
\end{equation}
\begin{equation}
    [w_{i1}, w_{i2}, ..., w_{iN}]=\mathrm{softmax}(\frac{\xvec_i \cdot \xvec_1}{\sqrt{d}},\, \frac{\xvec_i \cdot \xvec_2}{\sqrt{d}},\,...\,,\frac{\xvec_i \cdot \xvec_N}{\sqrt{d}}) \label{eq:softmax_operation}
\end{equation}

Attention operation allows us to attend relevant vectors in the sequence by assigning higher weights and discard irrelevant vectors using lower weights.
For a given vector, say $\xvec_{i}$, if all the weights/similarities ($[w_{i1}, w_{i2}, ..., w_{iN}]$) are in a narrow range, it implies that all the vectors in $\Xmat$ are equally relevant to $\xvec_{i}$.
This could happen if all the vectors in the sequence belong to the same class.
In this case, the attention operation acts as, effectively, an averaging operation instead of a weighted average.
Consequently, we may not be exploiting transformer abilities to the maximum level.
Based on this insight, we hypothesize that input sequences with less categorical variety hinder transformer model performance. 
Equivalently, training data with input sequences containing diverse emotion classes provide better performance compared to sequences with less emotional diversity.

We validate our hypothesis by proposing a data augmentation scheme, referred to as \textit{DiverseCatAugment} (\textit{DCA}), which improves the diversity of categories/emotions in the input sequences.
We apply \textit{DCA} on conversations as well as isolated utterances.
When applying \textit{DCA} to conversations, we choose two conversations and concatenate them for model training.
For example, assume one conversation is filled with angry for most of the time and another with happy category.
Concatenation of the two conversations results in a sequence with both angry and happy.
It is easy to see that the concatenated conversations have a more diverse composition of emotions.
According to the \textit{DCA} hypothesis, proposed transformer models perform better if input sequences have diverse categories.

\section{Experimental setup}
\label{sec:experimental_setup}

\subsection{Dataset}
\label{sec:dataset}
We performed CER on the widely used IEMOCAP dataset, which contains 150 dyadic conversations between 5 female and 5 male speakers.
Each conversation is set up between one male and one female, and are approximately 5 min long.
The scripts and topics for spontaneous conversations were selected to elicit emotions.
Even though only 5 emotions -- \textit{Angry}, \textit{Frustration}, \textit{Happy}, \textit{Neutral}, and \textit{Sad} -- are targeted for elicitation, more emotions albeit less frequently are found in the annotation process.
In this work, we used only the most frequent emotions, -- \textit{Angry}, \textit{Frustration}, \textit{Happy}, \textit{Neutral}, and \textit{Sad} -- for classification.
We merged \textit{Excitation} emotion with \textit{Happy} as is commonly done for this dataset.
To facilitate comparison between models trained with isolated utterances and conversations, we discarded segments in the conversations which have labels other than the considered emotions.
As the dataset contains only 5 sessions, we use 3 sessions for training, 1 for development, and 1 for testing. 
We performed a 5-fold cross-validation (leave-one-out-session) and report the weighted average f1-score (micro-f1) score for our experiments.

\subsection{DCA implementation}
We implement \textit{DCA} augmentation during the formation of the batch for model training.
We first choose a batch of 6 conversations and pick a sequence of length 1024 from each of the conversations. 
Then, we randomly pair each conversation with one of the other 5 conversations to form a sequence of length 2048 for model training.
We note that to maximize \textit{DCA} utility, conversations with distinct emotions should be selected for concatenation but as we train the model for 100 epochs, the model sees a fairly high number of sequences with diverse emotions.
\textit{DCA} on conversations produces sequences with conversational context preserved for most of the sequence and adds a bit of random context.
When applying \textit{DCA} on isolated utterances, we concatenate multiple isolated utterances until we obtain 2048 length sequences.
We choose isolated utterances for concatenation randomly to result in a sequence with diverse emotions expressed by multiple speakers.
\textit{DCA} on isolated utterances results in sequences similar to conversations but without conversational context.

\subsection{Impact of the context}
\label{subsubsec:method_impact_context}

To gain insights into the model capabilities and importance of context, we compare the transformer model with ResNet-34 and BiLSTM using 4 types of training data: 
\begin{enumerate}
    \item \textit{Isolated utterances} (no context)
    \item \textit{Conversations} (original conversational context)
    \item \textit{DCA Isolated utterances} (random context) 
    \item \textit{DCA Conversations} (original conversational context + random context)
\end{enumerate}
    
We evaluate all the models on conversations.
We compare models trained with \textit{Isolated utterances} (no context) and \textit{Conversations} to understand the impact of conversational context on the CER performance. 
To further improve the performance, we employ \textit{DCA} on \textit{Isolated utterances} and \textit{Conversations}.

Based on the \textit{DCA} method hypothesis, we expect \textit{DCA Conversations} and \textit{DCA Isolated utterances} to perform better than \textit{Conversations} and \textit{Isolated utterances} respectively.
Also, as context could help to disambiguate emotions, we expect models trained on conversational data (2$^{nd}$ and 4$^{th}$ types) to perform better than models trained on isolated utterances data (1$^{st}$ and 3$^{rd}$ types).
The performance of models trained with isolated utterances enables us to answer the question of ``how well can we perform CER without access to the conversational data?''.
The answer to this question is important because most of the current datasets have only isolated utterances and a lot of past research efforts focused on them.

\section{Results}
\label{sec:results}

\subsection{Results with \textit{DCA} augmentation and context}

Table~\ref{tab:conversations_vs_isolated} shows the results with \textit{DCA} augmentation and context.
The first and second rows, denoted with \textit{Isolated} and \textit{Conversations}, show the results of models trained with isolated utterances and conversations.
We can observe that models trained on isolated utterances perform worse than the models trained on conversations suggesting the importance of context.
BiLSTM seems to predict just a little better than chance when trained on isolated utterances.
The impact of conversational context on the BiLSTM model is comparatively higher than ResNet and transformer.
Among the architectures, the transformer model outperformed ResNet and BiLSTM in every case with the best performance of 42\% when trained on conversations.

\begin{figure}[h]
    \centering
    \includegraphics[width=0.48\textwidth, scale=0.5]{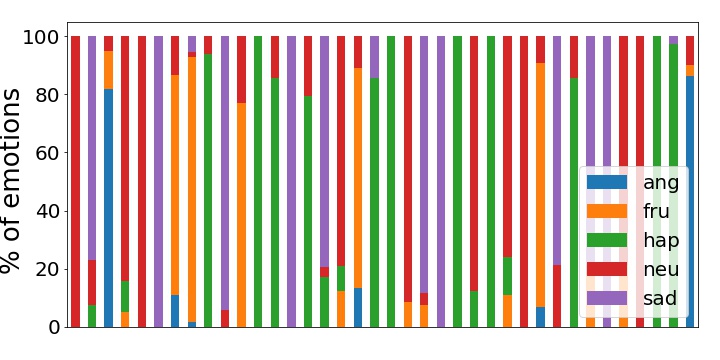}
    \caption{Proportion of emotions in a subset of 38 IEMOCAP dataset conversations (25\% of the dataset). These conversations have one emotion occurring for more than 75\% of the conversation. Each bar corresponds to one conversation}
    \label{fig:emotional_inertia_iemocap}
\end{figure}

Models trained with \textit{DCA} augmentation are denoted with \textit{DCA Isolated utterances} and \textit{DCA Conversations}.
We can observe that along with the transformer model, ResNet and BiLSTM also perform better with \textit{DCA} augmentation on isolated utterances suggesting that emotional variety in the training sequences helps to discriminate emotions well.
On isolated utterances, ResNet and BiLSTM models perform 3.9\% and 13.1\% absolute better with \textit{DCA} augmentation.
However, they perform worse in comparison to \textit{Conversations} suggesting that original conversational context is more important than categorical/emotional variety in the training sequences.
Interestingly, the transformer model trained with \textit{DCA Isolated utterances} performs better than \textit{Conversations}.
Upon further investigation into the conversations, we found that many conversations are dominated by a single emotion.
Fig.~\ref{fig:emotional_inertia_iemocap} shows proportions of emotions for a subset of 38 conversations (25\% of the dataset) in the IEMOCAP dataset.
Each bar represents the proportion of emotions in a single conversation. 
We can observe that these conversations have only one emotion dominating for more than 75\% of the conversation time.
In literature, this phenomenon is referred to as emotional inertia~\cite{kuppens2010emotional} which states humans naturally tend to resist changing emotions.

Emotional inertia in the conversations explains the better performance with \textit{DCA Isolated utterances} compared to \textit{Conversations} even though the latter has conversational context.
It also implies that emotional variety in the training sequences is important for the transformer model confirming the \textit{DCA} augmentation hypothesis.
Better (3.3\% absolute) performance with \textit{DCA Conversations} over \textit{Conversations} further strengthens the \textit{DCA} augmentation hypothesis.

Overall, we observed that training the models with random context is better than no context. 
Access to the conversational context further improved our models' performance.
Transformer model trained with conversations and \textit{DCA} augmentation performed best with a micro-f1 of 45.3\%.

\begin{table}\caption{Effect of context on the CER performance (micro-f1). Conv. context means the original conversational context; \textit{DCA Isolated utterances} -- \textit{DCA} augmentation on isolated utterances; \textit{DCA Conversations} --  \textit{DCA} augmentation on conversations}
    \label{tab:conversations_vs_isolated}
    \centering
    \resizebox{\columnwidth}{!}{
    \begin{tabular}{l|l|lll}
    \toprule
     Training data type  & Context type &  ResNet & BiLSTM & Transformer \\
    \midrule
    \textit{Isolated utterances} &  No   & 34.3 &  27.1    &  39.1 \\
    \textit{Conversations} &  Conv.  &  39.2  & 41.6     & 42.0 \\
    \midrule
    \textit{DCA Isolated utterances} & Random   &   38.1    &    40.2  & 42.7     \\
    \textit{DCA Conversations} &  Random+Conv.    &    37.5  & 41.6  &  45.3 \\
    \bottomrule
    \end{tabular}}
    
\end{table}

\subsection{Results with ResNet+Transformer and its analysis per emotion}
\label{subsec:resnet+transformer_results}

\begin{table}\caption{Results of joint ResNet and transformer training. \textit{DCA} on conversations is employed for model training}
    \label{tab:resnet_transformer}
    \centering
    \begin{tabular}{c|c}
    \toprule
    Model &  micro-f1 \\
    \midrule
     Transformer      &  45.3 \\
    ResNet+Transformer  & 49.8 \\
    \bottomrule
    \end{tabular}
    
\end{table}

\begin{figure}[h]
    \centering
    \includegraphics[width=0.45\textwidth, scale=0.5]{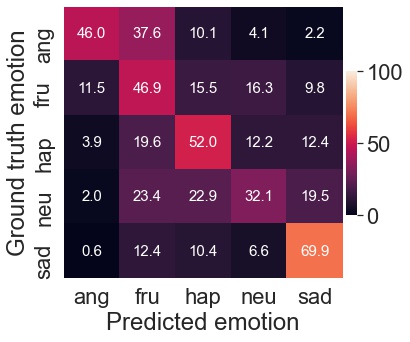}
    \caption{Confusion matrix of ResNet+Transformer model}
    \label{fig:confusion_matrix_resnet_transformer}
\end{figure}

Table~\ref{tab:resnet_transformer} compares the results of the ResNet+Tranformer model with only the transformer model.
We can observe 4.5\% absolute improvement in CER performance suggesting that processing with convolutional layers helps.
For this model, we employed \textit{DCA} augmentation on conversations as it yielded the best results.
To understand our model errors, we show an analysis of our model's row-normalized confusion matrix in Fig.~\ref{fig:confusion_matrix_resnet_transformer}.
We can observe that our model is confusing \textit{Angry} with \textit{Frustration} 37.6\% of the frames and \textit{Neutral} with other emotions 67.9\% of the frames.
\textit{Angry} and \textit{Frustration} seem much more similar to each other than to any other emotion in the label set, hence we wondered whether there could be some confusion between them for annotators too.
Looking at inter-annotator agreement, we found that when the annotation of each crowd-sourced worker is matched against their majority-voted annotation, \textit{Angry} is found to be confused with \textit{Frustration} 17\% and \textit{Frustration} with \textit{Angry} 11\% of total segments~\cite{busso2008iemocap} which are significantly high compared to any other emotion.
These confusion rates in the ground-truth annotations would explain our model's confusion to some extent.

\begin{figure}[h]
    \centering
    \includegraphics[width=0.45\textwidth, scale=0.5]{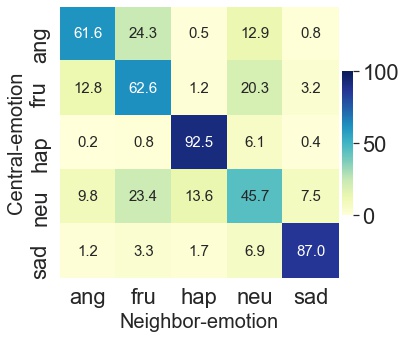}
    \caption{Probabilities of trigrams of the form (neighbor-emotion, central-emotion, neighbor-emotion). The labels ang, fru, hap, neu, and \textit{sad} stand for \textit{Angry}, \textit{Frustration}, \textit{Happy}, \textit{Neutral}, and \textit{Sad} respectively. }
    \label{fig:trigram_prob_dataset}
\end{figure}

To understand the confusion of \textit{Neutral} emotion with others, we investigated the trigram probabilities of emotions in the conversations.
Fig.~\ref{fig:trigram_prob_dataset} shows dataset statistics for a subset of trigrams of the form (neighbor-emotion, central-emotion, neighbor-emotion).
These statistics are computed from sequences of turn/segment emotions.
Each row is normalized for analysis purposes.
If central-emotion is equal to neighbor-emotion then we call the trigram as homogeneous, and heterogeneous otherwise.
From Fig.~\ref{fig:trigram_prob_dataset}, we can observe that the majority of trigrams are homogeneous (diagonal values) except when the central-emotion is \textit{Neutral}.
Approximately 54.3\% (100\%-45.7\%) of the trigrams are heterogeneous for \textit{Neutral} emotion compared to 38.5\%, 37.5\%, 7.5\%, and 13.1\% for \textit{Angry}, \textit{Frustration}, \textit{Happy}, and \textit{Sad} respectively.
We speculate that the heterogeneous nature of \textit{Neutral} in this dataset could be one reason why our model confuses with other emotions more often -- it simply "prefers" to recognize longer contiguous segments with a single emotion, mislabeling \textit{Neutral} in the process.
This observation is consistent with the hypothesis posed by~\cite{pappagari2020copypaste} that neutral utterances are perceived as emotional when presented in the context of another emotional utterance.
However, whether this behavior is because of the dataset characteristics or the acoustic characteristics of \textit{Neutral} emotion warrants further analysis which we plan to address in future work.

\subsection{Results with interlocutor-aware ResNet+Transformer} 

Table~\ref{tab:interlocutor_for_training} presents the results of models trained with and without interlocutor information.
We can observe a 3.2\% absolute improvement in micro-f1 with the addition of interlocutor information to the model training.
With the interlocutor information, the model could be able to distinguish the emotions of interlocutors from the past and future to predict the emotion of the current location.
The class-wise analysis further revealed that a significant contribution to the overall gain is from \textit{Neutral} and \textit{Angry} emotions with 20\% and 15\% relative improvements.
We suspect the higher improvements specifically for \textit{Neutral} and \textit{Angry} emotions are because of their higher percentage of heterogeneous trigrams (refer to the discussion in Section~\ref{subsec:resnet+transformer_results}).
As heterogeneous trigrams are formed from a sequence of three turns, by definition, the central turn has different emotion and speaker compared to neighbor turns.
We think the interlocutor-aware model can better distinguish central turn's emotion from neighbor turns' emotions as it has access to the speaker diarization information.

\begin{table}\caption{Influence of interlocutor information on the performance of ResNet+Transformer model. \textit{DCA} on conversations is employed for model training}
    \label{tab:interlocutor_for_training}
    \centering
    \resizebox{\columnwidth}{!}{
    \begin{tabular}{@{}c|c c c c c c@{}}
    \toprule
    \begin{tabular}[c]{@{}c@{}} Interlocutor \\ Index \\ Embedding\end{tabular}  & \textit{Angry} & \textit{Frustration} & \textit{Happy} & \textit{Neutral} & \textit{Sad} & Micro-f1\\
    \midrule
    No     & 47.1 &    47.9  & 50.1 &  33.2  &   67.6 & 49.8 \\
    Yes &  54.2 & 48.5 & 54.7 & 39.9 & 68.8 & 53.0 \\
    \bottomrule
    \end{tabular}
    }
\end{table}

\section{Conclusions and future work}
\label{sec:conclusion}
In this work, we presented transformer-based models for conversational emotion recognition (CER).
Our analysis on the impact of context showed that models trained with random conversational context perform better on conversations than those trained without context from other speakers.
We found that less diversity of emotions/categories in the input sequences limits the transformer model performance.
Our proposed data augmentation scheme which aims to improve diversity has helped to discriminate the emotions better.
Conversational context and diversity of emotions provided the best results when using transformers.
The proposed transformer-based approaches always outperformed the baseline architectures ResNet-34 and BiLSTM.
We presented a model combining ResNet-34 and transformer architecture to exploit local and global context, that provides better results than the model based on transformer only.
We have shown that a model with interlocutor information improves the CER performance.

Except for the interlocutor-aware transformer model, models presented in this work do not assume a single emotion per turn and can deal with multi-party conversations without requiring speaker-diarized recordings.
The interlocutor-aware transformer model requires speaker-diarized recording and the number of speakers in a conversation at test time can not be more than the maximum number of speakers seen in a training conversation.
In the future, we plan to propose models to overcome this limitation.
In this work, we evaluated the proposed methods on the IEMOCAP corpus.
Some of the shortcomings of this corpus are its limited number of speakers and its collection in controlled settings. 
We plan to evaluate our models on more spontaneous conversations data with more speakers such as MELD~\cite{poria2019meld}.

\bibliographystyle{IEEEtran}

\bibliography{main}

\end{document}